\documentclass[runningheads, a4paper]{llncs}
\usepackage[T1]{fontenc}
\usepackage{soul}
\usepackage[dvipsnames]{xcolor}
\usepackage{booktabs}
\usepackage{multirow}
\usepackage{float} 

\usepackage{graphicx}
\usepackage{amsmath} 
\usepackage{amssymb}

\usepackage{microtype}

\begin{document}

\title{A Benchmark for Hallucination Detection in VLMs for Gastrointestinal Endoscopy}

\titlerunning{Hallucination Detection in VLMs for GI Endoscopy}

\author{
Aminu Lawal\inst{1}\thanks{Corresponding author.}
\and Niyoj Oli\inst{2}
\and Sachin Acharya\inst{2}
\and Prashnna Gyawali\inst{3}
\and Maria Carmen Romano\inst{1}
\and Binod Bhattarai\inst{1}
}

\authorrunning{A. Lawal et al.}
\institute{
University of Aberdeen, Aberdeen, United Kingdom \\
\email{\{a.lawal.25, m.romano, binod.battarai\}@abdn.ac.uk}
\and
Nepal Applied Mathematics and Informatics Institute for Research, Lalitpur, Nepal
\email{\{oliniyoj, sachinacharya44\}@gmail.com}
\and
West Virginia University, Morgantown, WV, United States
\email{prashnna.gyawali@mail.wvu.edu}
}
\maketitle              
\setcounter{footnote}{0}

\begin{abstract}
Vision-language models (VLMs) are prone to hallucination, which remains a major barrier to their safe deployment in clinical practice. To date, most hallucination detection methods have been evaluated on radiology benchmarks such as MIMIC-CXR and VQA-RAD, while gastrointestinal (GI) endoscopy remains largely underexplored. In this paper, we benchmark nine hallucination detection methods on the Gut-VLM dataset, a GI diagnostic Visual Question Answering (VQA)  dataset with 4,392 test VQA pairs, across five VLMs (MedGemma-4B, MedGemma-27B, LLaVA-Med-7B, LLaVA-v1.6-7B, and Lingshu-32B). The methods span three categories: black-box methods (RadFlag, SelfCheckGPT-NLI), gray-box methods (AvgProb, AvgEnt, MaxProb, MaxEnt, Semantic Entropy, and VASE), and a white-box method (ReXTrust). Our results show that ReXTrust, a white-box method, achieves the highest AUC across all five models, outperforming the strongest alternative method on each VLM by a statistically significant margin (paired permutation test, p\,<\,0.001 in all cases), reaching a peak AUC of 93.0 on MedGemma-4B. White-box hidden-state access provides a consistent advantage of 19.5 AUC points on average (range: 9.5--33.5), with ReXTrust maintaining strong performance even on LLaVA-v1.6-7B (AUC 79.9), where black-box methods and clustering-based gray-box methods collapse to near-chance performance. Among non-white-box methods, token-level gray-box statistics (MaxEnt, MaxProb) are the strongest alternatives, outperforming both clustering-based gray-box methods (Semantic Entropy, VASE) and black-box approaches on average. We further identify confident confabulation, a failure mode in which models hallucinate with high inter-sample consistency or high token-level probability, as a systemic failure for both consistency and uncertainty-based methods.


\end{abstract}

\section{Introduction}
Advances in large vision-language models (VLMs) have recently driven substantial progress in medical reasoning, such as medical report generation \cite{Mota2024A,radflag}.  These models demonstrate the ability to jointly interpret visual clinical data and generate natural language responses to diagnostic queries \cite{liao2025vase}, positioning them as promising tools for clinical decision support. However, despite these capabilities, hallucination, the generation of outputs that are \emph{linguistically plausible yet inconsistent} with the visual evidence, remains a critical limitation. In high-stakes clinical settings, where such errors may manifest as missed detections (e.g., overlooked polyps) or spurious findings (e.g., falsely reported lesions), hallucinations pose a significant threat to the reliability and safety requirements of real-world deployment.

Consequently, substantial effort has been directed toward detecting and mitigating hallucinations in medical VLMs.
Existing hallucination detection approaches vary in the level of model access, with trade-offs in accuracy, computational cost, and deployment accessibility \cite{rextrust,selfcheckgpt,radflag}. Despite this progress, most benchmarking has focused on radiology datasets such as MIMIC-CXR \cite{johnson2019mimiccxrjpg}, SLAKE \cite{liu2021slake}, PathVQA \cite{he2020pathvqa}, and VQA-RAD \cite{lau2018vqarad}, while gastrointestinal (GI) endoscopy has received far less attention. 
Unlike radiology, GI endoscopy presents a distinct imaging environment \cite{Arnold2020Global}, marked by heterogeneous image quality, variable lighting conditions, and organ-dependent visual patterns. In addition, clinically relevant GI findings, including polyps, ulcerative colitis, oesophagitis, and anatomical landmarks, differ substantially in appearance from those encountered in radiological imaging.
As a result, it remains unclear whether methods validated on radiology datasets transfer effectively to this domain.
In this work, we address this gap by systematically evaluating hallucination-detection methods for GI endoscopy VQA.

We evaluate nine hallucination-detection methods, covering multiple methodological paradigms, within a unified experimental framework, allowing direct comparison under consistent conditions. We further analyze detection behavior in endoscopic imagery to characterize the effects of domain shift on method reliability and to derive practical insights for the deployment of hallucination-aware clinical VLM pipelines. Our contributions are summarized as follows:

\begin{enumerate}
    \item We conduct a comprehensive evaluation of nine hallucination detection methods across five VLMs, revealing that methods validated on radiology benchmarks fail to transfer reliably, particularly on general-purpose VLMs where all non-white-box methods fall to near-chance performance. 

    \item We identify confident confabulation as a systemic failure mode that simultaneously defeats both consistency-based and uncertainty-based detection, and show that white-box hidden-state access (ReXTrust) outperforms the strongest non-white-box method on each of the five evaluated VLMs by a statistically significant margin ($p < 0.001$ in all cases)
\end{enumerate}

\begin{figure}[h!]
    \centering
      \includegraphics[width=\linewidth]{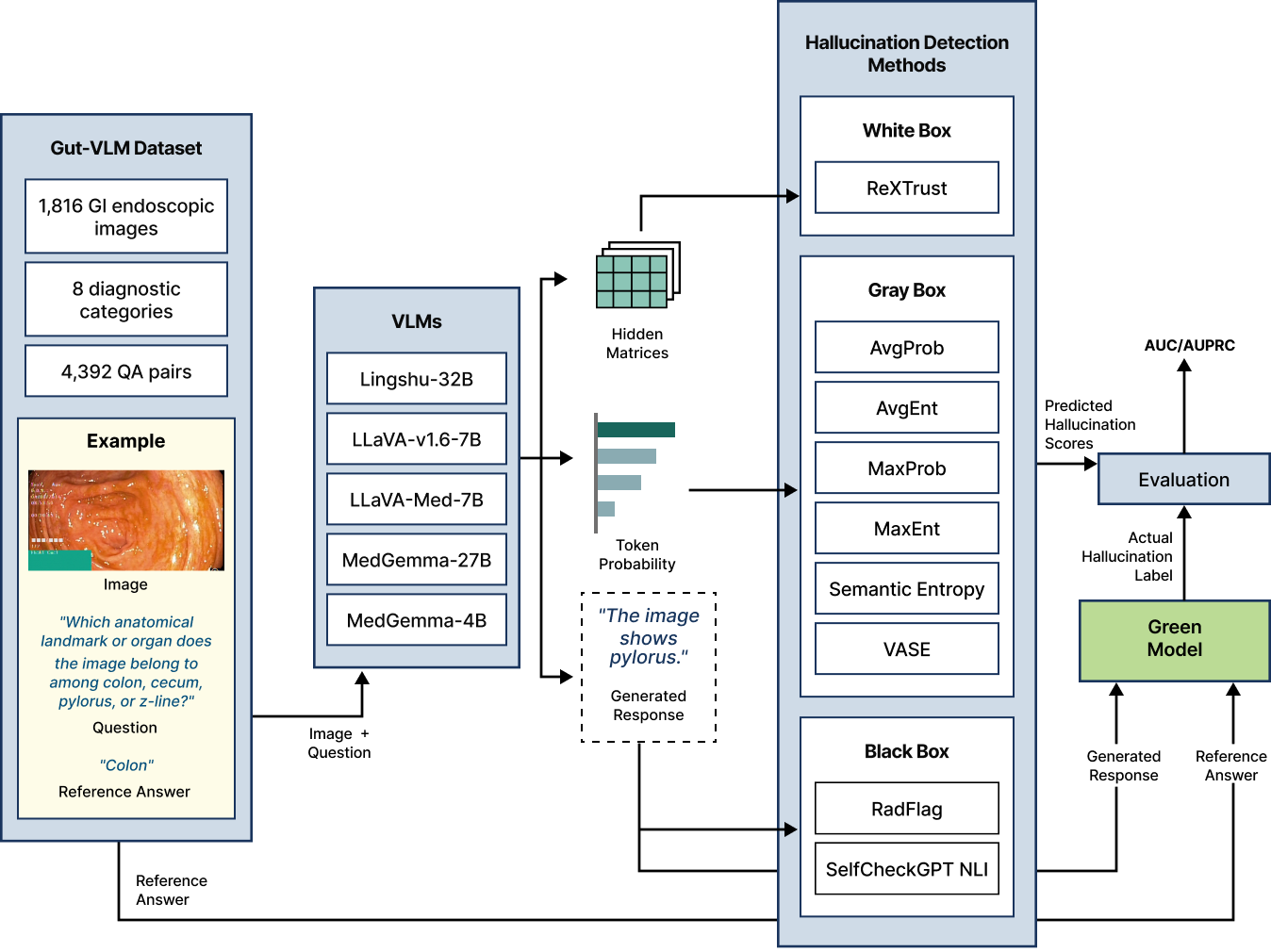}
    \caption{ Overview of the benchmark pipeline. An image-question pair from Gut-VLM is passed to five VLMs, which produce hidden states, token probabilities, and generated responses utilized by nine hallucination detection methods across three access categories. Actual hallucination labels are derived independently by the GREEN model, which compares the generated response against the expert-verified reference answer. Detection performance is evaluated using AUC and AUPRC.}
\label{fig:pipeline}
\end{figure}

\section{Methodology}

 In this section, we present a detailed overview of our evaluation framework as illustrated in Fig. \ref{fig:pipeline}. We first describe the dataset used in our experiments, followed by the VLMs evaluated in this work. We then introduce the categories of hallucination detection algorithms evaluated in this benchmark.

\subsection{Dataset: Gut-VLM} 
\begin{figure}[h!]
  \centering
  \setlength{\tabcolsep}{2pt}
  \begin{tabular}{cccc}
    \includegraphics[width=0.23\linewidth]{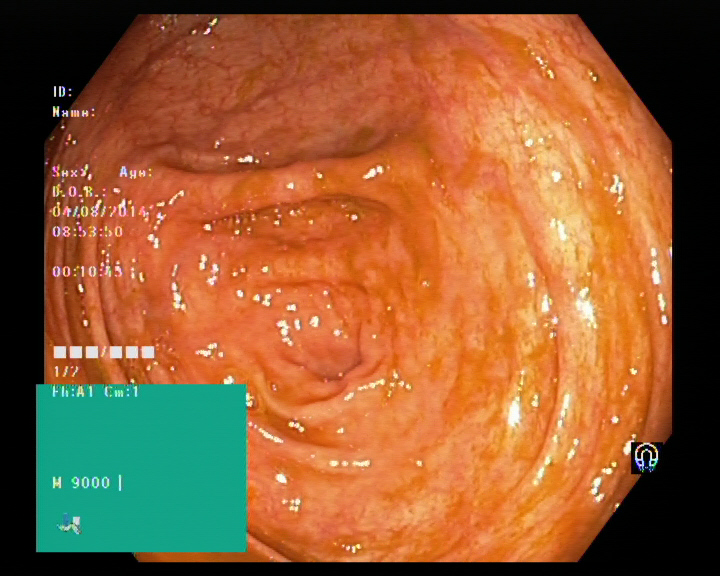} &
    \includegraphics[width=0.23\linewidth]{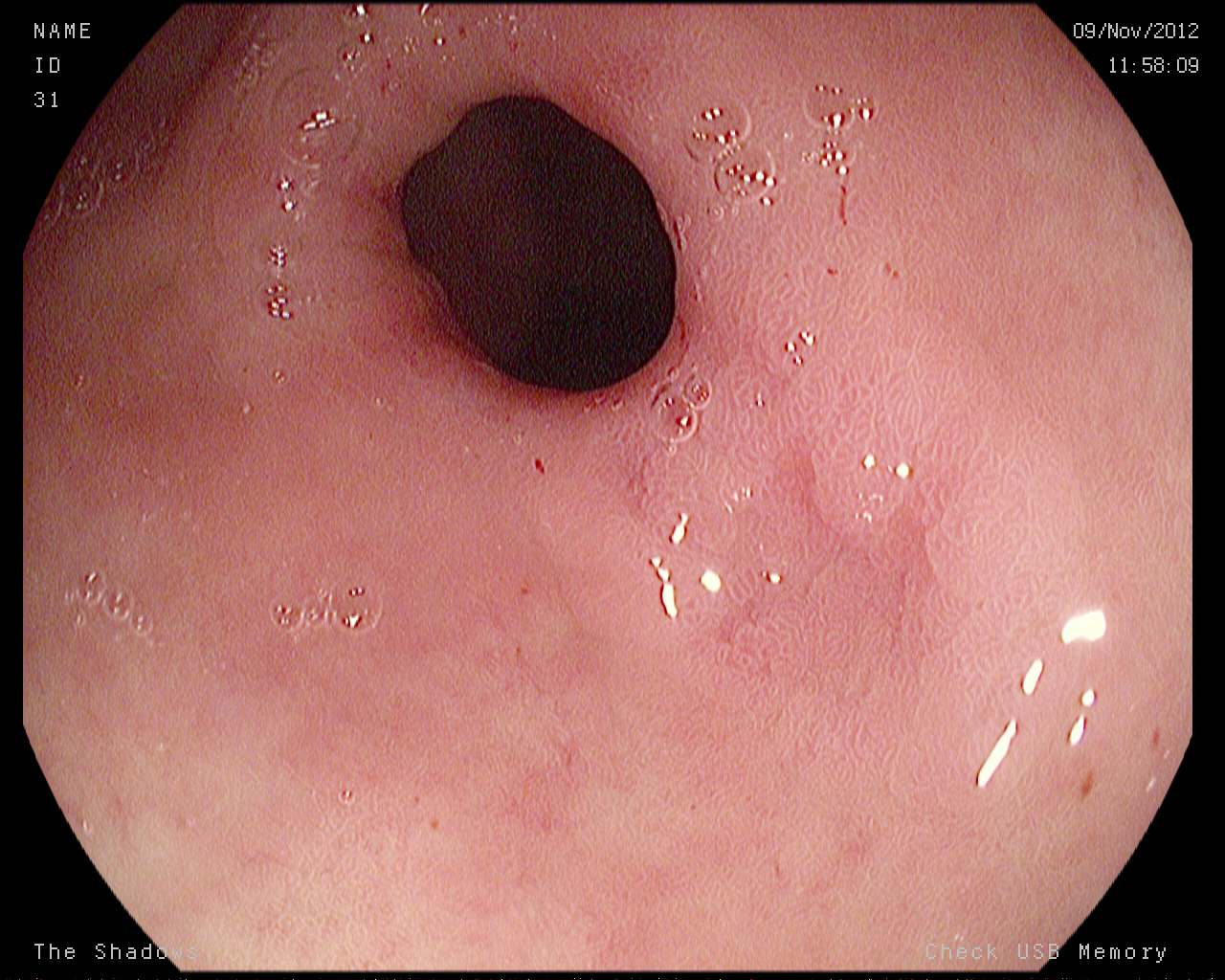} &
    \includegraphics[width=0.23\linewidth]{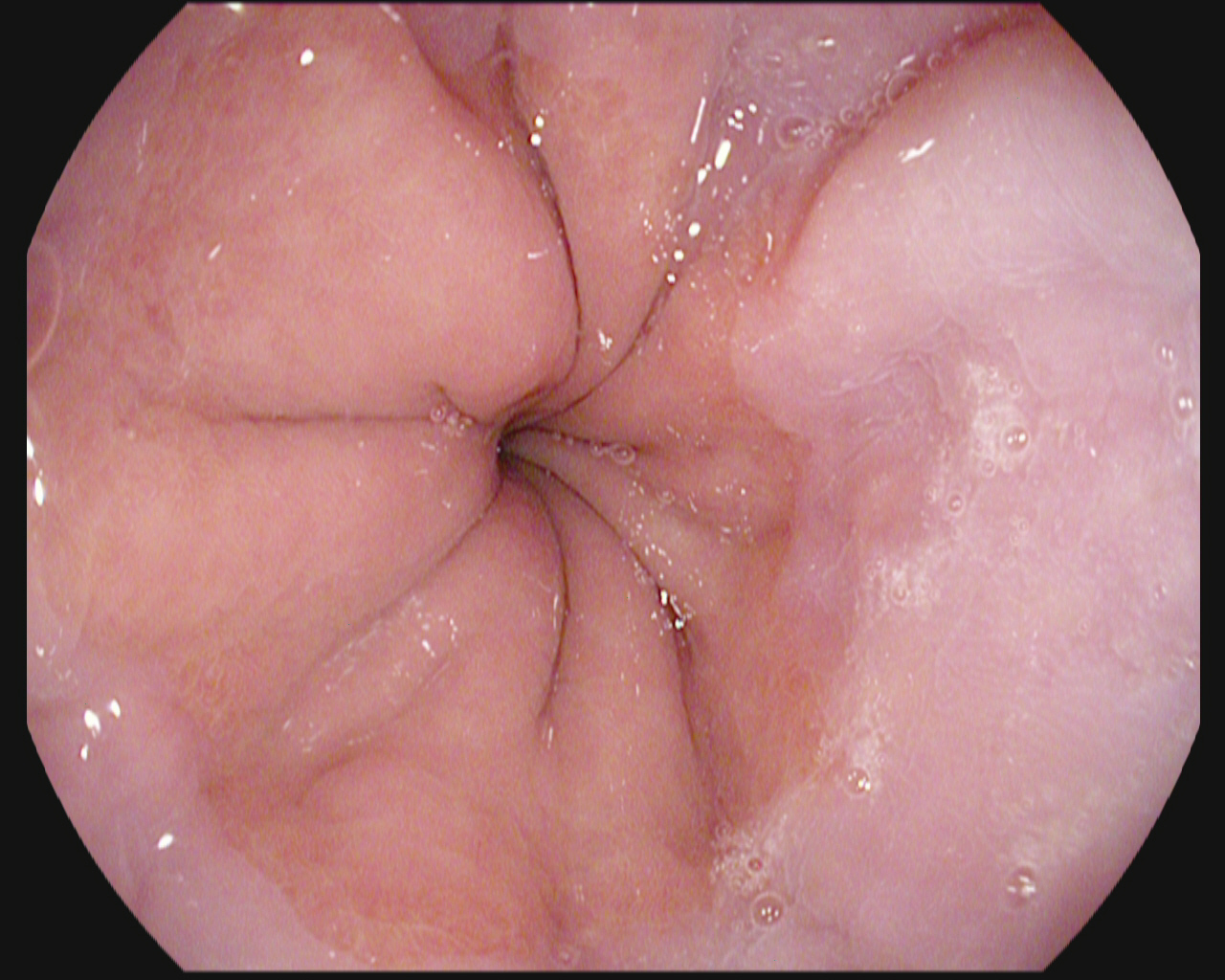} &
    \includegraphics[width=0.23\linewidth]{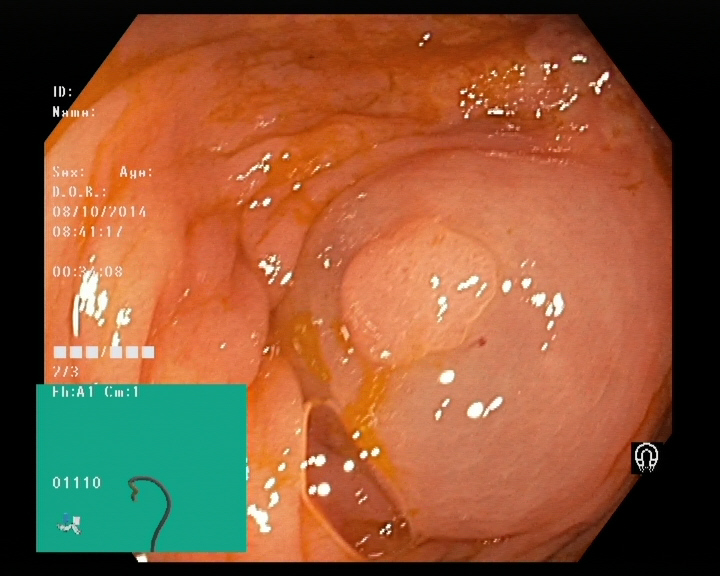} \\
    \small Cecum & \small Pylorus & \small Z-line & \small Polyps \\[4pt]
    \includegraphics[width=0.23\linewidth]{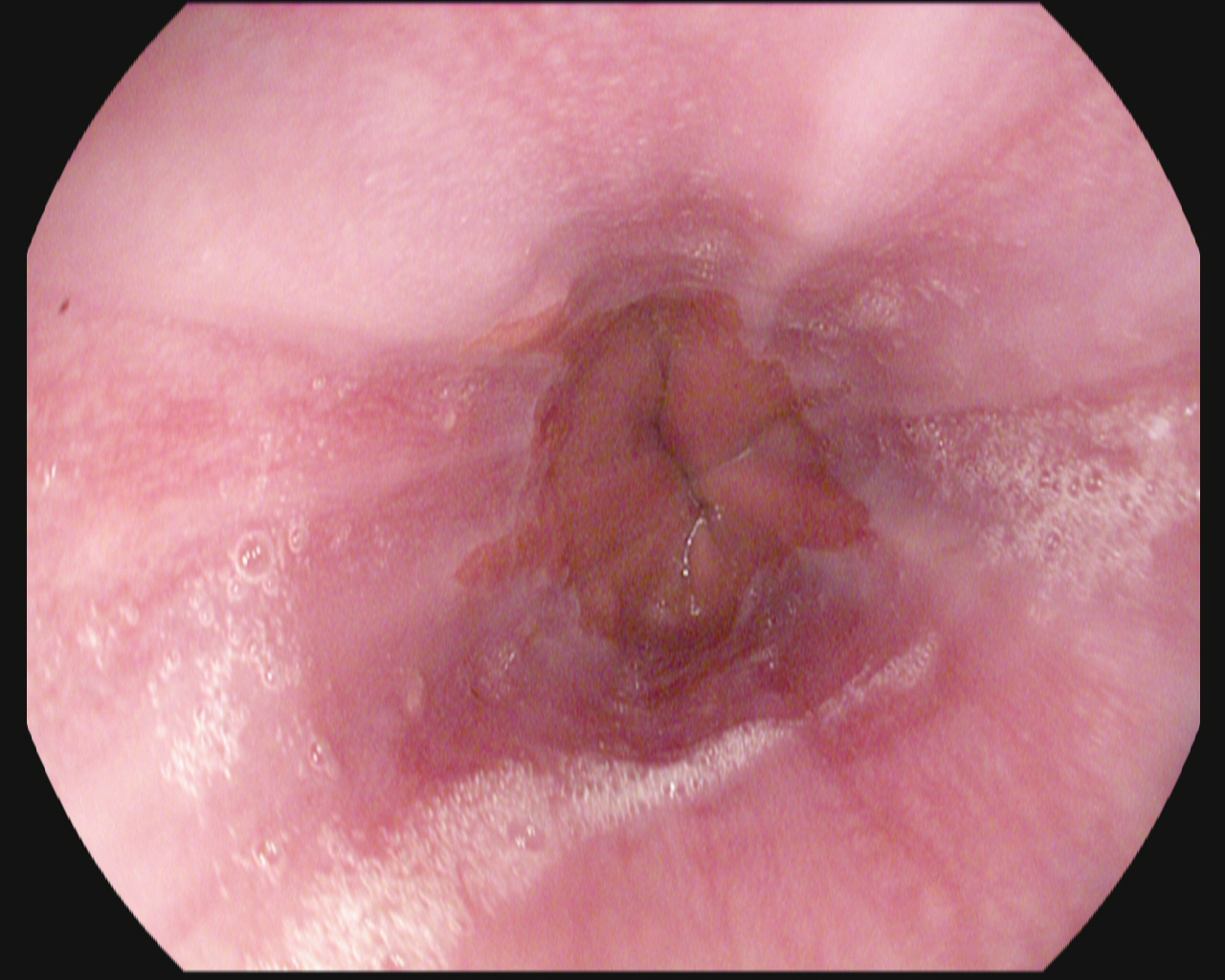} &
    \includegraphics[width=0.23\linewidth]{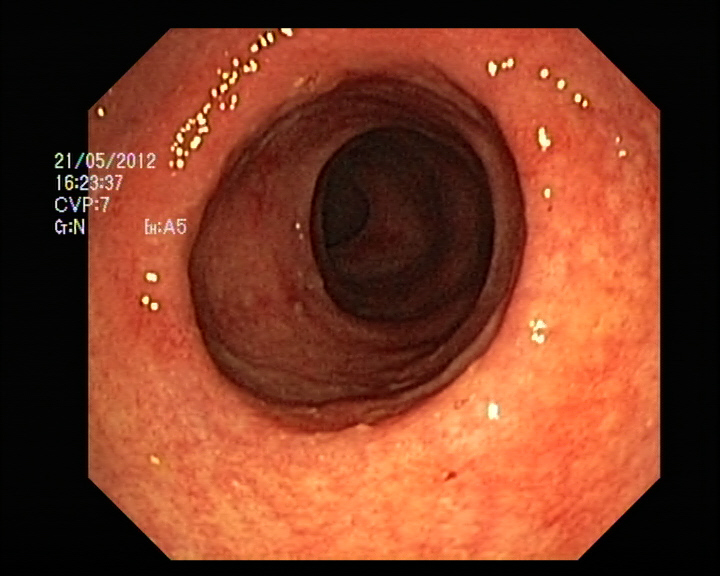} &
    \includegraphics[width=0.23\linewidth]{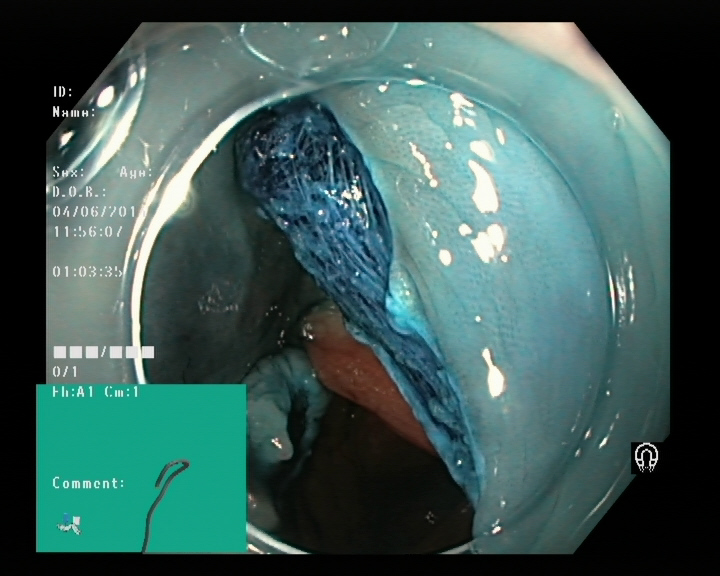} &
    \includegraphics[width=0.23\linewidth]{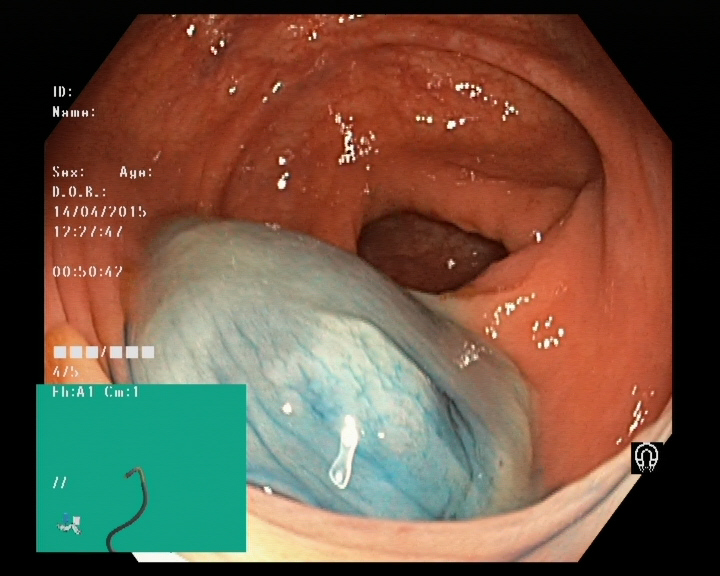} \\
    \small Oesophagitis & \small Ulcerative Colitis & \small Dyed-resected & \small Dyed-lifted \\
  \end{tabular}
  \caption{Representative endoscopic images from the eight diagnostic categories 
  in the Gut-VLM benchmark~\cite{khanal2025hallucination}. \textit{Top row}: anatomical landmarks 
  (normal findings). \textit{Bottom row}: pathological findings. 
  }
  \label{fig:gut-VLM_samples}
\end{figure}

We select Gut-VLM \cite{khanal2025hallucination} \footnote{\url{https://github.com/bhattarailab/Hallucination-Aware-VLM}} as it is a publicly available dataset for GI endoscopy designed specifically for hallucination detection, built on Kvasir-v2 \cite{pogorelov2017kvasir} with expert-verified reference answers used as ground truth for hallucination labeling. The dataset comprises 1,816 endoscopic images spanning eight diagnostic categories, including three anatomical landmarks (Cecum, Pylorus, Z-line) representing normal findings, and five pathological findings (Polyps, Oesophagitis, Ulcerative Colitis, Dyed-resected-margins, and Dyed-lifted-polyps) as shown in Fig.~\ref{fig:gut-VLM_samples}. Following the stratified partitioning established by \cite{khanal2025hallucination}, we utilize a 20\% hold-out per category, resulting in a training set of 1,450 images (17,400 QA pairs) and a test set of 366 images (4,392 QA pairs).

This VQA framework enables a more rigorous examination of model reliability; specifically, each image is paired with 12 structured visual questions designed to probe diverse clinical reasoning dimensions. By evaluating multiple query types grounded in the same visual input, this setting facilitates a fine-grained assessment of hallucination behavior, ensuring that model responses are truly rooted in visual evidence rather than language bias.

\subsection{Vision-Language Models}
We evaluate five VLMs selected to cover a range of model scales, training approaches, architectures, and domain orientations.
MedGemma-4B \cite{sellergren2025medgemma} is a 4-billion parameter multimodal model from Google, built on the Gemma 3 language backbone with a SigLIP vision encoder and trained on diverse medical image-text data spanning radiology, dermatology, and pathology. MedGemma-27B \cite{sellergren2025medgemma} is the 27B variant of the same architecture, enabling direct comparison of scale within the family. LLaVA-Med-7B \cite {li2023llavamed} uses a CLIP ViT-L/14 vision encoder with a 7B LLaMA-based language model fine-tuned on approximately 600,000 biomedical image-text instruction pairs from PubMed Central. Lingshu-32B \cite{team2025lingshu} is a 32-billion parameter generalist medical VLM developed by DAMO Academy (Alibaba Group), representing the current frontier of medical specialist VLMs. LLaVA-v1.6-7B (Mistral) \cite {liu2023visual} is based on the LLaVA-NeXT framework with a Mistral-7B language model and improved high-resolution visual processing. It was not trained on medical data, which makes it a useful reference for assessing whether medical-domain training affects hallucination detectability.

\subsection{Hallucination Detection Algorithms}

\noindent\textbf{Problem Formulation.}
Let $\mathcal{M}$ be a VLM that processes image $x \in \mathcal{X}$ and question $q \in \mathcal{Q}$. The inputs $(x,q)$ are tokenized and decoded autoregressively, producing token-level probability distributions $\mathbf{P}_\theta^{*} = \{P_\theta(y_t \mid x, q, y_{<t})\}_{t=1}^{T}$ and response $y = \mathcal{M}(x,q)$. The hidden-state matrix at layer $l = \lfloor n_{\text{layers}}/2\rfloor$ is then extracted as $\mathbf{H}_\theta^{(l)}(x,q,y) \in \mathbb{R}^{(T+1)\times D}$, where row~0 is the final prompt token (prefill anchor), rows $1\ldots T$ are the $T$ generated tokens, and $D$ is the model's hidden dimension. A response $y$ is \textit{hallucinated} if it contains claims not grounded in $(x,q)$. The hallucination detection problem is to learn a scoring function $h: \mathcal{X} \times \mathcal{Q} \times \mathcal{U} \rightarrow [0,1]$ such that $h(x, q, u)$ is a scalar hallucination score, with
\begin{equation}
     \hat{y}^{\text{hall}} = \mathbf{1}\left[h(x, q, u) > \tau\right]
\end{equation}
for threshold $\tau$. The three method families below differ in which component $u \in \mathcal{U}$ they access: sampled outputs $\mathcal{S}$ only (black-box), token distributions $\mathbf{P}_\theta^{*}$ (gray-box), or hidden states $\mathbf{H}_\theta^{(l)}$ (white-box).

\noindent\textbf{Black-Box Methods:}
Black-box methods utilize only the generated response (including stochastic samples), without accessing internal states, logits, or parameters \cite{selfcheckgpt,radflag,factselfcheck}. We generate $N$ stochastic responses $\mathcal{S} = \{y_1, \ldots, y_N\}$ from $\mathcal{M}(x, q)$. Setting $u:= \mathcal{S}$, the hallucination detector is then defined as:
\begin{equation}
     h_{\text{BB}}(x, q, u) = h_{\text{BB}}(x,q, \mathcal{S})
\end{equation}
where no model internals are accessed by $h_{\text{BB}}$. RadFlag \cite{radflag} is a black-box method designed specifically for medical VLMs. It generates multiple stochastic samples and scores the primary response by measuring its entailment-based consistency with those samples. SelfCheckGPT-NLI \cite{selfcheckgpt} uses DeBERTa-v3-large \cite{debertav3} to assess whether each sentence in the primary response is contradicted by any remaining sample, flagging frequently contradicted sentences as hallucinated.

\noindent\textbf{Gray-Box Methods:} Gray-box methods utilize token-level probability distributions, either from a single forward pass \cite{shorinwa2025surveyuqllms} or in combination with stochastic or visually-augmented samples, without requiring access to weights or hidden states. 
Let $\mathbf{P}_\theta^{*} = \{P_\theta(y_t \mid x, q, y_{<t})\}_{t=1}^{T}$ denote the per-token output distributions from the single deterministic (greedy) forward pass, with $J \leq T$ valid (non-special) tokens. Setting $u:= \mathbf{P}_\theta^{*}$, the hallucination detector for the token-level gray-box methods is:
\begin{equation}
    h_{\text{GB}}(x, q, u) = h_{\text{GB}}\!\left(x,q,\mathbf{P}_\theta^{*}\right).
\end{equation}

We evaluate four token-level statistics following \cite{li2024referencefree}, all derived from this single greedy pass:
\begin{align}
\text{AvgProb} &= -\frac{1}{J}\sum_{t=1}^{J}\log P_\theta(y_t \mid x, q, y_{<t}) \label{eq:avgprob} \\
\text{MaxProb} &= \max_{t \in \{1,\dots,J\}}\bigl(-\log P_\theta(y_t \mid x, q, y_{<t})\bigr) \label{eq:maxprob} \\
\text{AvgEnt}  &= \frac{1}{J}\sum_{t=1}^{J} H\!\left(P_\theta(\cdot \mid x, q, y_{<t})\right) \label{eq:avgent} \\
\text{MaxEnt}  &= \max_{t \in \{1,\dots,J\}} H\!\left(P_\theta(\cdot \mid x, q, y_{<t})\right) \label{eq:maxent}
\end{align}
where $H(\cdot)$ denotes Shannon entropy. All four are uncertainty scores: higher values indicate greater uncertainty and predict hallucination. \textit{AvgProb} and \textit{MaxProb} are computed as the mean and maximum negative log-probability (NLL) of generated tokens, respectively. \textit{AvgEnt} and \textit{MaxEnt} measure mean and maximum per-token entropy of the full output distribution at each generation step. All four are reference-free and require no additional stochastic samples.

Unlike the four token-level methods above, Semantic Entropy (SE) \cite{semanticentropy} and  VASE \cite{liao2025vase} require stochastic responses $\mathcal{S} = \{y_1,\ldots,y_N\}$ sampled from $\mathcal{M}(x,q)$ at temperature $> 0$. SE clusters $\mathcal{S}$ into semantic equivalence classes and computes entropy over the resulting cluster distribution using model likelihoods. 
VASE additionally generates augmented-image samples $\mathcal{S}_{\text{aug}}$ from $\mathcal{M}(x',q)$ under weak visual perturbations $x'$.

\noindent\textbf{White-Box Methods:}
White-box methods access the model's internal hidden-state representations \cite{rextrust}, requiring open-source model access but enabling the strongest detection signal. Using $\mathbf{H}_\theta^{(l)}(x,q,y)$ as defined above, and setting $u:= \mathbf{H}_\theta^{(l)}(x, q, y)$, the detector is:
\begin{equation}
    h_{\text{WB}}(x, q, u) = h_{\text{WB}}\!\left(x,q,\mathbf{H}_\theta^{(l)}(x, q, y)\right),
\end{equation}
achieving the strongest performance but requiring open-source model access. We adapted the ReXTrust \cite{rextrust} framework for hallucination detection in GI endoscopy VQA for this study.

\section{Experiments}
\subsection{Implementation Details}
All experiments were conducted on a single NVIDIA A100 GPU across all evaluations. All models were loaded with 8-bit weight quantization ( INT8 via BitsAndBytes) with bfloat16 activations and responses capped at 128 tokens.  
All nine methods share a single unified inference pipeline that produces one deterministic (greedy) primary response and one GREEN-derived hallucination label per sample, ensuring that all methods are evaluated on identical inputs and ground-truth labels. 
For SelfCheckGPT-NLI, RadFlag, Semantic Entropy, and VASE, $N = 10$ independent responses per question at temperature 1.0, top-$p$ 0.9 were sampled, ensuring a fair and consistent comparison across all stochastic methods. 
The four token-level statistics (AvgProb, AvgEnt, MaxProb, MaxEnt) were computed from the single greedy forward pass with logit access following Equations~(\ref{eq:avgprob})--(\ref{eq:maxent}). Semantic Entropy clustering used agglomerative clustering with DeBERTa-v3-large NLI entailment scores as the similarity measure, following \cite{semanticentropy}. ReXTrust was trained on the Gut-VLM training split. For each training sample, hidden states were extracted from the middle transformer layer (layer index $\lfloor n_\text{layers}/2 \rfloor$, architecture-specific: layer 9 for MedGemma-4B, 14 for MedGemma-27B, 16 for LLaVA-Med-7B and LLaVA-v1.6-7B, and 32 for Lingshu-32B). 
The classifier was trained on up to 2,400 samples per model using 5-fold cross-validation with GroupKFold partitioning, grouping all 12 QA pairs from the same image into the same fold to prevent data leakage. The five-fold models were then weight-averaged to produce the final classifier. A WeightedRandomSampler was used during training to address class imbalance, with a cosine-annealing learning rate schedule and gradient clipping (max norm 1.0). Code will be made publicly available upon paper acceptance.

\noindent \textbf{Hallucination labeling:}
Hallucination labels for all nine evaluated methods were derived using the GREEN model \cite {ostmeier2024green}, which scores generated VLM responses against expert ground-truth answers from Gut-VLM. GREEN evaluates clinical accuracy across multiple dimensions and assigns a score between 0 and 1. Responses receiving a GREEN score below 1.0 indicate the presence of at least one clinical inaccuracy (hallucinated response). Responses with a GREEN score of 1.0 were labeled as non-hallucinated, while all other scores were labeled as hallucinated. This binarization threshold was applied consistently across all methods. This threshold is the natural binary split in GREEN's design: a score of exactly 1.0 denotes full factual alignment across all evaluated clinical dimensions, while any score below 1.0 indicates that at least one clinical error was detected. The criterion is intentionally conservative; the sensitivity of our labeling to alternative thresholds is acknowledged in Section~\ref{sec:limitations}. For ReXTrust, the same criterion was used to derive the binary training labels required for classifier training.

\subsection{Evaluation Metrics}
Area Under the ROC Curve (AUC) measures the probability that the detection method assigns a higher hallucination score to a hallucinated QA-pair than a non-hallucinated one across all thresholds. The Area Under the Precision-Recall Curve (AUPRC) is particularly informative for the positive (hallucinated) class but is more sensitive to prevalence than AUC. 
All nine methods are evaluated within the single unified inference pipeline that produces one greedy primary response and one GREEN-derived hallucination label per sample. Consequently, the hallucination rate is identical across all methods for a given model (Table~\ref{tab:auc_auprc}), and AUPRC is directly comparable across methods within a model, though it remains sensitive to per-model hallucination prevalence. AUC is adopted as the primary metric precisely because it is threshold-independent and less sensitive to prevalence differences than AUPRC.

\begin{table}[t]
\centering
\small
\caption{Hallucination prevalence and AUC / AUPRC (\%) per method and model. All nine methods share a unified pipeline with identical GREEN-derived labels per model. Bold = best AUC per column; underline = second best.}
\resizebox{\textwidth}{!}{%
\begin{tabular}{llccccc}
\hline
\textbf{Category} & \textbf{Method} & MedGemma-4B & LLaVA-Med-7B & LLaVA-v1.6-7B & MedGemma-27B & Lingshu-32B \\
\hline
\multicolumn{2}{l}{\textit{Hallucination rate \%}} & 76.5 & 57.3 & 74.0 & 61.1 & 44.1  \\
\hline
\multirow{2}{*}{Black-Box}
 & RadFlag \cite{radflag}          & 37.03 / 73.05 & 47.10 / 56.92 & 45.64 / 72.12 & 54.66 / 61.76 & 45.28 / 39.34 \\
 & SelfCheckGPT-NLI \cite{selfcheckgpt} & 50.12 / 77.09 & \ul{76.30} / 76.28 & 55.47 / 80.59 & 56.24 / 60.33 & 42.76 / 37.16 \\
\hline
\multirow{6}{*}{Gray-Box}
 & AvgProb \cite{li2024referencefree} & 52.36 / 75.61 & 48.90 / 60.56 & 56.06 / 78.92 & 73.40 / 74.45 & 72.65 / 71.05 \\
 & AvgEnt \cite{li2024referencefree}  & 51.88 / 76.36 & 44.43 / 58.25 & \ul{56.32} / 79.27 & 73.67 / 77.08 & 74.63 / 74.83 \\
 & MaxProb \cite{li2024referencefree} & 59.20 / 85.59 & 55.30 / 66.23 & 52.41 / 75.33 & \ul{80.93} / 81.99 & 70.15 / 60.36 \\
 & MaxEnt \cite{li2024referencefree}  & \ul{59.46} / 86.02 & 61.60 / 68.40 & 55.32 / 78.32 & 80.85 / 81.72 & \ul{74.70} / 62.02 \\
 & Semantic Entropy \cite{semanticentropy} & 39.32 / 75.43 & 44.72 / 55.44 & 49.00 / 75.56 & 57.83 / 66.91 & 52.89 / 45.28 \\
 & VASE \cite{liao2025vase}            & 40.23 / 75.60 & 44.87 / 56.86 & 48.38 / 75.14 & 58.42 / 68.52 & 49.86 / 45.60 \\
\hline
White-Box & ReXTrust \cite{rextrust} & \textbf{92.99} / 97.17 & \textbf{90.46} / 90.71 & \textbf{79.91} / 92.21 & \textbf{90.39} / 92.50 & \textbf{91.43} / 88.89 \\
\hline
\end{tabular}}
\label{tab:auc_auprc}

\end{table}

\subsection{Quantitative Analysis}
Table ~\ref{tab:auc_auprc} reports the AUC and AUPRC for all nine detection methods across the five VLMs on the Gut-VLM test set. The most consistent pattern is the advantage of white-box access.

\begin{figure*}[h]
    \centering
    \includegraphics[width=\textwidth]{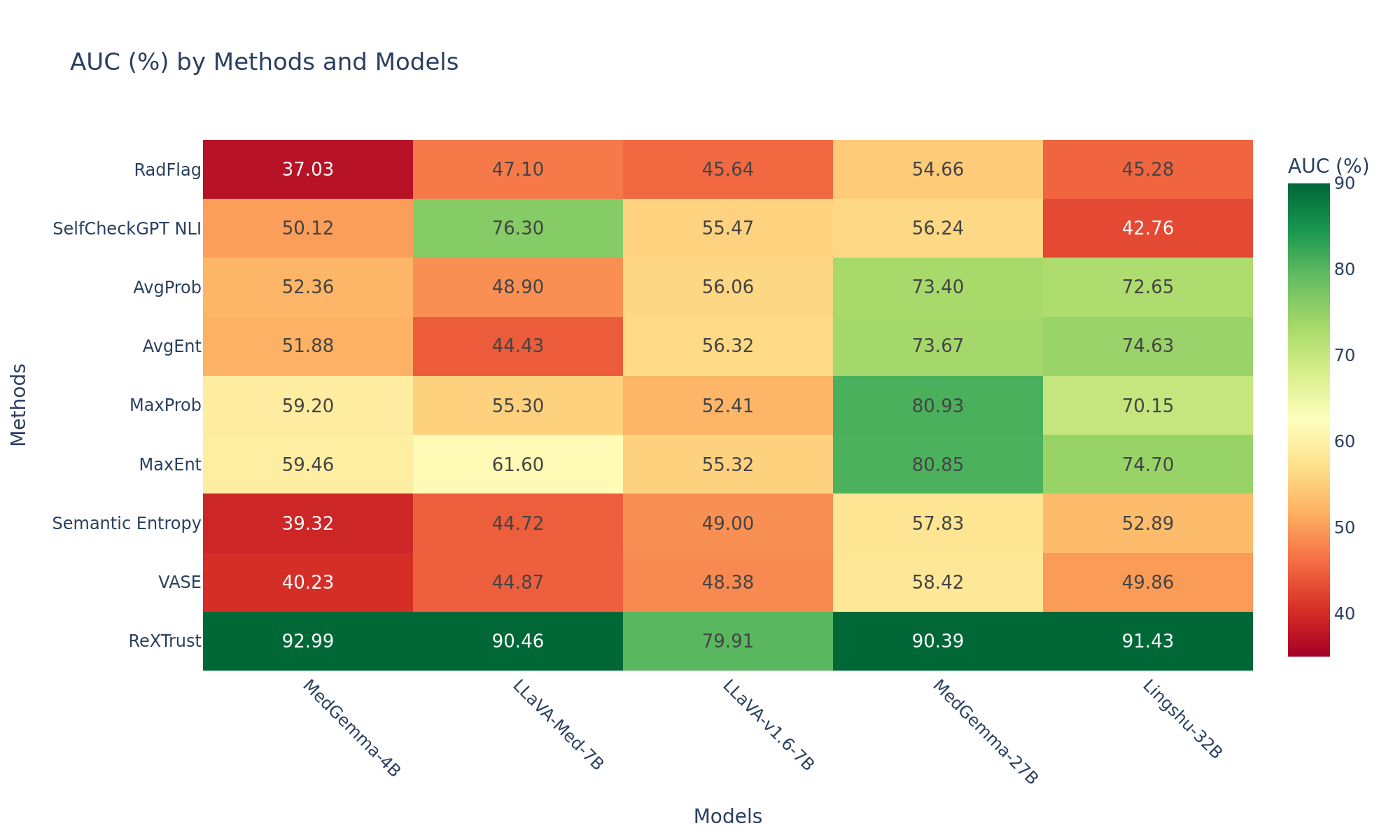}
    \caption{AUC (\%) of hallucination detection methods across five VLMs 
    on the Gut-VLM dataset. Higher values (green) indicate better detection 
    performance, while lower values (red) indicate poor performance.}
    \label{fig:auc_heatmap}
\end{figure*}

ReXTrust achieves the highest AUC across all five models (Table~\ref{tab:auc_auprc}, Fig.~\ref{fig:auc_heatmap}). Among non-white-box methods, the four token-level gray-box statistics (AvgProb, AvgEnt, MaxProb, MaxEnt) constitute the second-best category, achieving mean AUCs of 60--66 across models. Their advantage is most pronounced on the larger medical VLMs: MaxProb and MaxEnt reach 80.93 and 80.85 on MedGemma-27B, approaching ReXTrust. Semantic Entropy and VASE perform substantially below the token-level methods, with mean AUCs of 48.7 and 48.4 respectively, comparable to RadFlag (mean AUC 45.9).

Black-box methods show the widest variance. SelfCheckGPT-NLI peaks at 76.3 on LLaVA-Med-7B but collapses to 42.8 on Lingshu-32B, while RadFlag falls below chance on MedGemma-4B (37.0). On LLaVA-v1.6-7B, the general-purpose model, black-box and clustering-based gray-box methods fall to near-chance performance (AUC 45.6--55.5), while token-level gray-box methods achieve 52.4--56.3. ReXTrust achieves 79.9, suggesting that hidden-state representations may retain discriminative signal even without medical-domain training.

A paired permutation test ($N=2,000$, paired at the QA-pair level) comparing ReXTrust against the strongest competing method per model confirms that the white-box advantage is statistically significant on all five models ($p\,<\,0.001$ in every case, remaining significant under Bonferroni correction for five simultaneous tests, $\alpha_{\text{adj}}=0.01$ ): MedGemma-4B ($\Delta$=+33.5 vs.\ MaxEnt), LLaVA-Med-7B ($\Delta$=+14.2 vs.\ SelfCheckGPT-NLI), LLaVA-v1.6-7B ($\Delta$=+23.6 vs.\ AvgEnt), MedGemma-27B ($\Delta$=+9.5 vs.\ MaxProb), and Lingshu-32B ($\Delta$=+16.7 vs.\ MaxEnt).

\subsection{Qualitative Analysis}
\subsubsection {Hallucinated vs. Non-Hallucinated Responses}
\begin{figure}[t]
    \centering
    \includegraphics[width=\textwidth]{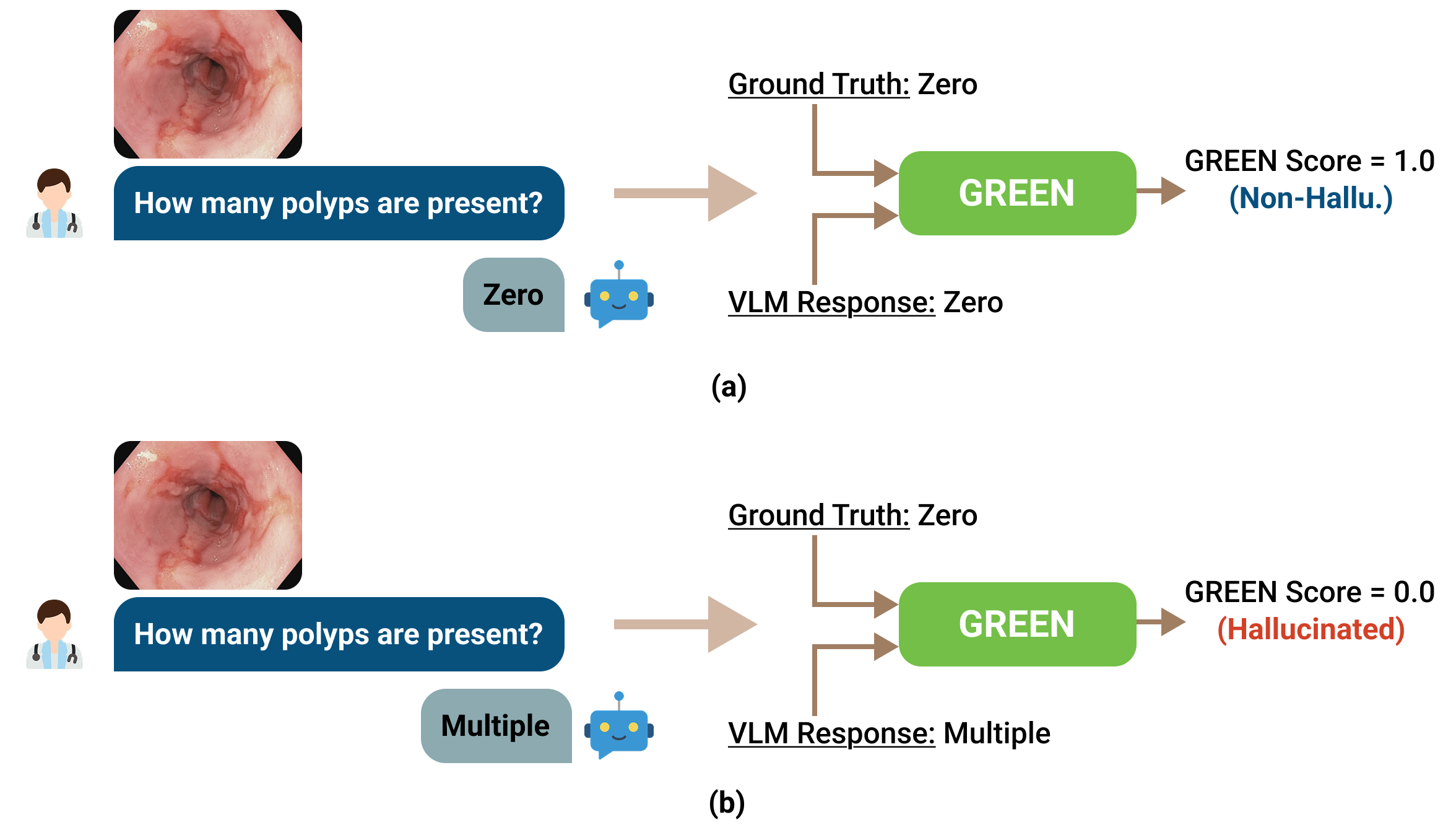}
    \caption{GREEN model-based hallucination labeling. The GREEN model scores generated answer and ground-truth answer pairs to produce hallucination labels. (a) A non-hallucinated example receiving a GREEN score of 1.0, indicating full factual alignment with the reference. (b) A hallucinated example receiving a GREEN score of 0.0, indicating a critical factual inconsistency with the ground truth.}
    \label{fig:hallucination-examples}
\end{figure}

Figure~\ref{fig:hallucination-examples} shows representative hallucinated and non-hallucinated response pairs illustrating how the GREEN model assigns scores to generated responses: a score of 1.0 indicates full factual alignment with the reference answer (non-hallucinated), while a score of 0.0 indicates a critical factual inconsistency with the ground truth (hallucinated).

\subsubsection{Confident Confabulation}

\begin{figure}
    \centering
    \includegraphics[width=\linewidth]{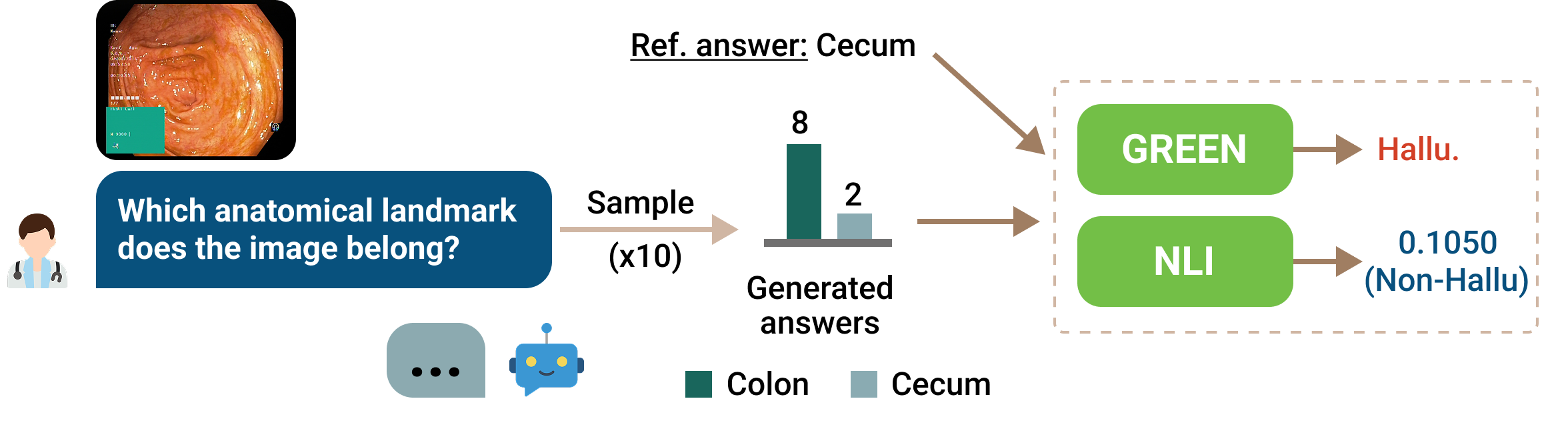}
    \caption{Confident confabulation in Lingshu-32B. The model predicts \textit{Colon} on 8 of 10 stochastic samples for an image depicting the \textit{Cecum}. The high inter-sample consistency yields a low SelfCheckGPT-NLI score (0.1050), causing the hallucination to be misclassified as non-hallucinated. Despite this, the GREEN model correctly identifies the factual error and assigns a hallucination label, highlighting the failure of consistency-based methods when a model is \textit{consistently wrong}.}
    \label{fig:se_limitation}
\end{figure}

Figure~\ref{fig:se_limitation} presents a concrete instance of the 
confident confabulation  \cite{lens}, a failure mode related to the confabulation described in \cite{semanticentropy}. For a Cecum image, Lingshu-32B produces \textit{Colon} across 8 of 10 stochastic samples, with only 
2 samples returning the correct answer \textit{Cecum}. Despite this 
being a hallucinated response, the SelfCheckGPT-NLI score is 0.1050,near the bottom of its scoring range. SelfCheckGPT-NLI scores measure average cross-sample contradiction: 0 indicates full consistency (no sentence contradicted by any sample) and 1 indicates complete contradiction. A score of 0.1050 reflects near-perfect agreement among stochastic samples, so the method ranks this response as highly non-hallucinated, causing the hallucination to be missed. 
See Section~\ref{sec:confab} for a full analysis of its structural implications.

\section{Discussion}
\subsection{Why Internal Access Matters}
The consistent advantage of ReXTrust comes from direct access to the VLMs hidden states \cite{rextrust}. Prior work indicates that during hallucination, internal attention patterns exhibit reduced alignment with visual input tokens \cite{Zhang2024DHCP}, while activation patterns in deeper layers diverge from those characteristic of grounded responses.  By contrast, token probabilities leveraged in gray-box methods capture only a partial projection of this internal information \cite{rextrust}. Across all five models, the white-box advantage averages 19.5 AUC points (range: 9.5--33.5, all $p\,<\,0.001$). The effect is most pronounced on LLaVA-v1.6-7B, where all non-white-box methods fall to near-chance (AUC 45.6--56.3) while ReXTrust achieves 79.9, demonstrating that hidden states retain discriminative hallucination signal even without medical-domain training. On LLaVA-Med-7B, the strongest competitor (SelfCheckGPT-NLI, AUC 76.3) remains 14.2 points below ReXTrust (90.5).

\subsection{Confident Confabulation: A Systemic Failure of Consistency and Uncertainty-Based Detection}
\label{sec:confab}
The instance in Fig. \ref{fig:se_limitation} is not isolated. Across the five evaluated models, 0.0--20.4\% of hallucinated samples exhibit confident confabulation (inter-sample agreement $\geq 0.8$), with the highest frequency on Lingshu-32B (20.4\%), followed by MedGemma-4B (15.3\%) and MedGemma-27B (11.6\%); overall, 8.9\% of all hallucinated samples across all models show this pattern. SelfCheckGPT-NLI underperforms on Lingshu-32B because it actively assigns \textit{lower} hallucination scores to wrong answers than to correct ones. The method relies on the assumption that hallucinations manifest as inconsistency across stochastic samples and facts reproduce consistently, while hallucinated content varies \cite{selfcheckgpt}. This holds for open-domain text generation, but it breaks for medical VQA where a model systematically generates the same wrong answer across all stochastic samples.

In our evaluation, Lingshu-32B generates short, confident responses (mean 2.96 words; 87.3\% of responses contain three words or fewer) 
regardless of factual accuracy. When the model says \textit{``Colon.''} consistently for an image that shows a cecum, NLI scores are low because all samples agree, and the response is classified as non-hallucinated despite being consistently wrong. This occurs when the model lacks GI-endoscopy-specific anatomical knowledge sufficient to recognise its error, so it hallucinates with confidence. Note that Lingshu-32B is a generalist medical VLM; despite broad medical pre-training, it is not specifically optimized for GI endoscopy anatomy, which accounts for this failure. 

This failure mode extends beyond consistency-based methods. Gray-box methods that derive hallucination signals from token-level probability distributions, such as AvgProb, AvgEnt, MaxProb, and MaxEnt, are equally vulnerable. When a model confabulates confidently, it assigns high probability mass to its hallucinated tokens, yielding low entropy and high average probability, which are precisely the signals these methods associate with \textit{non-hallucination}. Confident confabulation, therefore, exploits a shared assumption across consistency-based and uncertainty-based detection methods that hallucinations correlate with either inter-sample disagreement or token-level uncertainty. When neither holds, detection collapses regardless of method category.

\subsection{Practical Implications for Deployment}
From a deployment standpoint, the three method categories impose very different operational constraints. White-box methods require open-source models and separate training data per VLM. When white-box access is unavailable, the token-level gray-box methods (MaxEnt and MaxProb) are the strongest alternatives, achieving mean AUCs of 66 and 64 respectively with no additional stochastic sampling. Semantic Entropy and VASE, despite requiring $N=10$ stochastic samples, underperform the single-pass token methods on all five models in the unified pipeline, suggesting their benefit is contingent on generation diversity that is reduced when model outputs are short and deterministic. For deployments using proprietary VLMs where logits are inaccessible, SelfCheckGPT-NLI is the most reliable black-box option (mean AUC 56.2). RadFlag (mean AUC 45.9) falls below chance on MedGemma-4B (37.0) and should not be used without prior validation on the target architecture. SelfCheckGPT-NLI should be preferred, with the caveat that it fails on models generating consistently short, confident responses (e.g., Lingshu-32B: AUC 42.8).

\subsection{Clinical Relevance and Deployment Readiness}
The AUC values reported here represent aggregate ranking performance across all thresholds, but clinical deployment requires a fixed operating point. For a human-in-the-loop workflow \cite{azaria}, where the detector flags responses for clinician review rather than autonomously accepting or rejecting them, high recall is prioritized over precision. It is preferable to flag a correct response unnecessarily than to miss a hallucination. At a recall-oriented threshold (80\% recall), ReXTrust is likely to operate with acceptable precision across all five models, where AUC consistently exceeds 79. Not all hallucinations carry equal clinical weight; polyp and dyed-resection margin errors are directly actionable and potentially harmful \cite{Mota2024A}, while anatomical landmark misidentifications are less immediately dangerous.

Computational cost is a further barrier. ReXTrust requires a dedicated hidden-state extraction pass, while black-box methods need $N = 10$ generations per input, translating to hours of additional compute per clinical session. 
Semantic Entropy and VASE, at $N = 10$ samples, impose less sampling overhead than black-box methods, though their mean AUCs of 48--49 fall short of the deployment threshold. 
Based on our results, we use AUC\,$\geq$\,75 as a heuristic reference point for comparative discussion, rather than a clinically validated deployment threshold. 
This threshold is met by ReXTrust on all five models (79.9--93.0), by MaxEnt and MaxProb on MedGemma-27B (80.9), and by SelfCheckGPT-NLI on LLaVA-Med-7B (76.3). No non-ReXTrust method meets this threshold on MedGemma-4B or LLaVA-v1.6-7B.

\subsection{Limitations}
\label{sec:limitations}
Our ReXTrust adaptation is trained on up to 2,400 samples per model from the Gut-VLM training split due to computational resource constraint. 
The results are evaluated on a single GI endoscopy dataset (Gut-VLM). We evaluate hallucination detection methods only on single-turn VQA, whereas multi-turn dialogue, where hallucinations may compound, is not covered. Ground-truth labels depend on the GREEN model, which may not capture all clinically significant errors, particularly borderline cases where a response is technically imprecise but clinically acceptable. We evaluated only one white-box method, ReXTrust, as most contemporary white-box approaches require access to the full attention matrices or hidden states. This becomes computationally prohibitive for VLMs, such as LLaVA-v1.6-7B and Lingshu-32B, which use a fixed patch size with a dynamic number of patches, resulting in approximately 2,400 image tokens per image and consequently very large attention matrix dimensions.

\section{Conclusion}

We presented a comprehensive benchmark of hallucination detection methods for VLMs applied to GI endoscopy, evaluating nine methods across three categories on the Gut-VLM dataset. 
The key findings are that ReXTrust (white-box) consistently achieves the highest detection accuracy across all five evaluated VLMs within this benchmark, reaching an AUC of 93.0 on MedGemma-4B and maintaining strong performance even on LLaVA-v1.6-7B (AUC 79.9), where black-box methods and clustering-based gray-box methods collapse to near-chance performance. The mean white-box advantage is 19.5 AUC points (median 16.7; range: 9.5--33.5), significant across all models ($p\,<\,0.001$). Among non-white-box methods, token-level gray-box statistics (MaxEnt, MaxProb) are the strongest alternatives, achieving mean AUCs of 66 and 64 and outperforming clustering-based gray-box methods (Semantic Entropy mean 48.7, VASE mean 48.4) and all black-box approaches. Confident confabulation~\cite{lens} exposes a shared vulnerability: when a model hallucinates with high inter-sample consistency and high token-level confidence, both consistency-based and uncertainty-based detectors fail.
 This failure is not captured by either NLI scoring or entropy alone. These findings suggest that hallucination detection strategies cannot simply be transferred from radiology benchmarks to GI endoscopy without architecture-specific validation. 

\subsubsection*{Acknowledgments}
This work is supported by the Petroleum Technology Development Fund (PTDF) of the Federal Republic of Nigeria under the PTDF-OSS Program with grant number PTDF/ED/OSS/PHD/AL/0010/24.

\subsubsection*{Disclosure of Interests}
The authors have no competing interests to declare that are relevant to the content of this article.

\bibliographystyle{splncs04}
\bibliography{refs}
\end{document}